\pdfoutput=1

\documentclass[11pt]{article}

\usepackage[preprint]{acl}

\usepackage{times}
\usepackage{latexsym}

\usepackage[T1]{fontenc}

\usepackage[utf8]{inputenc}
\usepackage{booktabs}
\usepackage{caption}
\usepackage{subcaption}
\usepackage{enumitem}
\usepackage{authblk}
\usepackage{caption}
\usepackage{multirow}
\usepackage{wrapfig}
\usepackage{natbib}
\setcitestyle{round}
\usepackage{amsmath}
\usepackage{comment}
\usepackage{microtype}

\usepackage{inconsolata}

\usepackage{graphicx}
\usepackage[T1]{fontenc}
\usepackage{tcolorbox}
\title{Reuse, Don't Retrain:  \\ A Recipe for Continued Pretraining of Language Models}

\author{
  \textbf{Jupinder Parmar\thanks{Correspondence to: \texttt{jupinderp@nvidia.com}}},
  \textbf{Sanjeev Satheesh},
  \textbf{Mostofa Patwary},
  \textbf{Mohammad Shoeybi},
  \textbf{Bryan Catanzaro}

\\
   NVIDIA
   
}

\begin{document}
\maketitle
\begin{abstract}
As language models have scaled both their number of parameters and pretraining dataset sizes, the computational cost for pretraining has become intractable except for the most well-resourced teams. This increasing cost makes it ever more important to be able to reuse a model after it has completed pretraining; allowing for a model's abilities to further improve without needing to train from scratch. In this work, we detail a set of guidelines that cover how to design efficacious data distributions and learning rate schedules for continued pretraining of language models. When applying these findings within a continued pretraining run on top of a well-trained 15B parameter model, we show an improvement of 9\% in average model accuracy compared to the baseline of continued training on the pretraining set. The resulting recipe provides a practical starting point with which to begin developing language models through reuse rather than retraining.
\end{abstract}




\section{Introduction}

Language modeling abilities have seen massive improvements over the past few years \citep{brown2020language, chowdhery2022palm, openai2024gpt4, geminiteam2024gemini}. While these advancements have enabled language models (LMs) to become highly-skilled conversational agents \citep{openai2024gpt4, claude32024, geminiteam2024gemini}, they have come with increased computational cost as pretraining has become ever more expensive due to both the number of model parameters \citep{rekateam2024reka, deepseekai2024deepseek} and pretraining dataset size \citep{touvron2023llama2, gemma24, parmar2024nemotron4} continuing to grow in scale. With new LMs that set state of the art accuracy being released on a frequent basis, LMs developed only a couple months back are becoming obsolete as their capabilities are no longer up to par. This leaves model developers with the choice of either pretraining new LMs from scratch or reusing their existing LMs and updating them with new information in order to match current best LM abilities. 

Due to the large computational cost that pretraining of modern LMs incurs, frequent complete retraining is intractable. This makes the reuse of already developed LMs via continued pretraining an attractive proposition. While most recent works \citep{ibrahim2024simple, jang2022continual, ke2023continual, yıldız2024investigating} have recommended guidelines for continued pretraining when adapting language models to new data domains or distribution shifts, intuition or recommendations on how to improve a model's general purpose abilities from a previously finalized checkpoint with continued pretraining have not been widely explored. In this paper, we focus on this under-studied  setting and identify strategies that allow for already trained LMs to improve upon areas of weakness without experiencing degradations in other capabilities. 


In our experiments, we start on top of a 15B parameter LM that has seen 8T tokens of pretraining data \citep{parmar2024nemotron4}. Experimenting with a well trained model of this scale ensures that our findings will be transferable to most settings and model sizes. We first identify the type of data distribution that should be used during continued pretraining and find that it is optimal to have two distributions, with the final one more heavily weighting data sources that relate to the abilities we want to improve in the model. Second, we determine what learning rate schedules enable the most efficient learning during continued pretraining and determine that the most performant one strikes a balance between magnitude of learning rate and steepness of decay. Lastly, we show how the learning rate value at which we switch between data distributions affects downstream accuracy and identify the point at which this switch should be made.

These findings culminate in a recipe that can be used to perform continued pretraining to improve the capabilities of an existing LM. We demonstrate that this recipe is beneficial at continued training scales from 100B to 1 trillion tokens, illustrating its flexibility and robustness to be used in a wide variety of settings. We hope that this recipe will allow for model providers to forgo the need to regularly retrain models from scratch as it makes it possible to reuse a trained model to attain improved capabilities.



\section{Related Works}


Continued training methods aim to take an already trained model and incorporate new data, adapt it for a given domain, or specialize it on a certain task \citep{rolnick2019experience, caccia2021online, lesort2022understanding, gupta2023continual, lin2024rho1}. The major challenge that arises during continued training is enabling a model to learn new information without forgetting previously attained knowledge or capabilities \citep{Robins1995CatastrophicFR, FRENCH1999128}. The learning rate schedule and data distribution used during continued training \citep{gupta2023continual, ibrahim2024simple, winata2023overcoming, scialom2022finetuned} have been shown to be particularly important in preventing such catastrophic forgetting. 

For LMs, one major setting of continued training has been to embed more recent knowledge into the model by using data collected at a date later than when the pretraining set was constructed \citep{jin2022lifelong, jang2022continual, jang2023temporalwiki, loureiro2022timelms, qin2022elle}. Results from these studies found that using experience replay \citep{chaudhry2019tiny} and knowledge distillation \citep{hinton2015distilling} are particularly effective. Continued training is also commonly used in LMs to adapt the model to data coming from a new domain \citep{ke2023continual, gururangan-etal-2020-dont, wu2024llama}. Many of these methods for domain adaptive continued training update a portion of the model's weights with the new data to ensure that previous knowledge is not lost. For instance, \citep{wu2024llama} does so via an expansion of the transformer blocks and only updating the newly added weights.

More related to the setting which we explore, several studies utilize continued pretraining to specialize a LM on a given task or domain \citep{zan2022cert, yadav2023exploring, ma2023ecomgptct, yang2024pllama, labrak2024biomistral}. Despite investigating effective strategies for continued pretraining, these studies differ from ours as they do not aim to improve the general capabilities of LMs, train for far fewer tokens, and use much smaller model sizes. The main study which offers a comparative setting to ours is \citep{ibrahim2024simple} which provides a recipe, based on learning rate schedule and example replay recommendations, for maintaining general purpose abilities during continued pretraining on data distribution shifts. Their experimental setting consists of a 10B parameter model that was pretrained for 300B tokens. Our study differs from \citep{ibrahim2024simple} as we aim to improve the general capabilities of the LM further, and in our experimental setting we perform continued pretraining for up to 1T tokens with a 15B parameter model that was pretrained on 8T tokens.



\section{Experimental Setup}

The continued pretraining process is as follows: a model is first pretrained, then a data distribution and learning rate schedule are chosen, a continued pretraining run takes place, and finally the, hopefully improved, model is returned. Before delving into the experiments that define the continued training recipe, we detail the datasets and model architecture that are used.






\subsection{Data Sources}

\subsubsection{Pretraining}

Our pretraining dataset consists of three different domains of data: 
English natural language data, multilingual natural language data, and source code data. Table \ref{tab:data_high_level} highlights the data sources that compose the pretraining set along with their respective token counts. In our English corpus, the Web Crawl data is sourced from Common Crawl (CC) snapshots while the remaining categories are comprised of high-quality sets. For instance, the miscellaneous category consists of BigScience ROOTS \citep{lachaux2020unsupervised}, Reddit, and Pile-Stories \citep{pile-dataset-2020}, the encyclopedia category contains Wikipedia and Stack Exchange, and scientific papers includes ArXiv and PubMed. 

The multilingual dataset consists of 53 languages with the majority of examples being drawn from CC snapshots, although a small portion comes from machine translation parallel corpora \citep{schwenk2019ccmatrix, el2019ccaligned}. Lastly, our source code data is drawn from permissively licensed GitHub repositories and totals over 43 languages. 


\begin{table}[!t]
  \begin{tabular}{llr}
   \toprule
    \textbf{Data type} & \textbf{Data source} & \textbf{Tokens (B)} \\
    \midrule
    \multirow{8}{*}{English} & 
    Web Crawl &  5,106 \\
    &  Misc.   & 179  \\ 
     & News  & 93    \\
     & Scientific Papers &  82 \\ 
    & Books   & 80  \\ 
    & Legal &  50   \\
    & Encyclopedia &  31   \\
    & Finance &  20   \\
    \midrule
    \multirow{2}{*}{Multilingual} \hspace*{-0.25cm}
    & Web crawl   & 2,229  \\
    & Parallel corpora  & 55  \\
    \midrule 
    \multirow{1}{*}{Source Code} &  GitHub & 583  \\
    \bottomrule
  \end{tabular}
  \caption{The pretraining data composition. Appendix \ref{sec:appendix_data_sources_multilingual}  and \ref{sec:appendix_data_sources_code} breakdown the multilingual and coding languages. }
  \label{tab:data_high_level}
\end{table}

We pretrain the model for 8T tokens. Given that current state of the art LMs are pretrained for trillions of tokens, we want to experiment on top of a pretrained model that is emblematic of the type of models which the continued pretraining recipe would be used for.


\subsubsection{Continued Pretraining}

As the most likely scenario in continued pretraining is that the available datasets are exactly those which made up the pretraining set, the vast majority of our continued training data blend is comprised of the pretraining data sources. The only new additional source of data is a set of question and answer (QA), alignment style examples. Such examples have been shown to better extract stored knowledge within LMs \citep{allenzhu2023physics}. This set of QA data totals 2.8B tokens and Table \ref{tab:qa-data} highlights the categories of types of QA examples.

\begin{table}[!h]
  \begin{tabular}{llr}
   \toprule
    \textbf{Data type} & \textbf{Data source} & \textbf{Tokens (B)} \\
    \midrule
    \multirow{4}{*}{QA} & 
      World Knowledge   & 1.13  \\ 
    & Reasoning &  0.92 \\
     & STEM  & 0.31    \\
     & Chat &  0.26 \\ 
     & Code &  0.19  \\
    \bottomrule
  \end{tabular}
  \caption{The five constituent categories of the QA, alignment style data.}
  \label{tab:qa-data}
\end{table}

\subsection{Model Architecture and Hyperparameters}

We experiment using a 15B parameter decoder-only transformer \citep{Vaswani+2017} LM with causal attention masks. It has 3.2 billion embedding parameters and 12.5 billion non-embedding parameters. Additional architectural specifications include: 32 transformer layers, a hidden size of 6144, 48 attention heads, Rotary Position Embeddings (RoPE) \citep{su2023roformer}, squared ReLU activations in the MLP layers, a SentencePiece \citep{kudo2018sentencepiece} tokenizer with a vocabulary size of 256k, no bias terms, and untied input-output embeddings. Additionally, we use grouped query attention (GQA) \citep{ainslie2023gqa} with 8 KV heads.



The model is pretrained with a sequence length of 4,096 and uses batch size rampup over the first 5\% of pretraining tokens, starting from a batch size of 384 and building up to one of 1,152. We use a cosine learning rate schedule, with warmup of 16B tokens, to decay from a  maximum learning rate (LR) of $\eta_{max} = 4.5e\text{-}4$ to $\eta_{min} = 4.5e\text{-}5$. We train using the AdamW \citep{loshchilov2019decoupled} optimizer with $\beta_1 = 0.9$, $\beta_2 = 0.95$, and a weight decay of 0.1. In continued pretraining, the only hyperparameter that is altered is the learning rate schedule.






 

\subsection{Evaluation}

We evaluate the model using a representative set of tasks to test its change in abilities across the English, multilingual, and coding domains. To assess English capabilities, we evaluate on the widely-used MMLU \citep{hendrycks2020measuring} and Hellaswag \citep{Zellers2019HellaSwagCA} benchmarks. MMLU measures the model's world
knowledge across 57 domains while Hellaswag assesses commonsense reasoning ability within natural language inference. For our multilingual evaluations, we use the Multilingual Grade School Mathematics (MGSM) \citep{shi2022language} benchmark and specifically report the average accuracy across the language subset of Spanish, Japanese, and Thai, as they represent a high, medium, and low resource language respectively. Lastly, to assess the model's coding capabilities we utilize the Python code generation task of HumanEval \citep{chen2021evaluating} with evaluations reported in the pass@1 \citep{kulal2019spoc} setting. In our results below, we report the average score across all four of these tasks with fully detailed evaluation scores shared in the Appendix.

\section{Continued Pretraining Recipe}

The experimental findings which constitute our continued pretraining recipe are shared below:

\begin{tcolorbox}[width=\linewidth, colback=white!95!black, title={Recipe}]

\begin{itemize}
  \item Start with a data distribution that is similar to the pretraining set but places larger weight on high quality sources before transitioning to a second distribution that incorporates QA data and upweights sources in areas of model weakness.
  
  \item The learning rate schedule should start from $\eta_{min}$ of the pretrained model and decay with cosine annealing to $\frac{\eta_{min}}{100}$.
  
  \item The switch between data distribution should occur at $\frac{\eta_{max}}{5}$ in the learning rate schedule.
  
\end{itemize}

\end{tcolorbox}

\section{Experiments}

The results of the pretrained base model are shown in Table \ref{tab:base_model_results}. The aim for our continuous training recipe will be to define steps that help maximally improve upon this benchmark. All detailed experiments perform continuous pretraining for 300B tokens. Additionally, we note that in our experiments we choose to load in the optimizer state from the pretrained model as we found that there was a negligible difference in evaluation accuracy when the optimizer state was loaded in or when initialized from scratch. Thus, we expect that whether eventual practitioners have the optimizer state of the pretrained model available or not, the resulting findings will hold.

\begin{table}[!h]
\centering
  \begin{tabular}{lc}
   \toprule
    \textbf{Model} & \textbf{Average Accuracy} \\
    \midrule
    Pretrained &  48.9  \\
    \bottomrule
  \end{tabular}
  \caption{Model accuracy after 8T tokens of pretraining. Per-task evaluations scores are shared in Table \ref{tab:base_model_results_detailed}, we find the model particularly struggles on tasks that assess STEM based reasoning capabilities. }
  \label{tab:base_model_results}
\end{table}


\subsection{Data Distribution}
\label{sec:data_distributions}

\begin{figure*}[t]
    \centering
    \includegraphics[width=\linewidth]{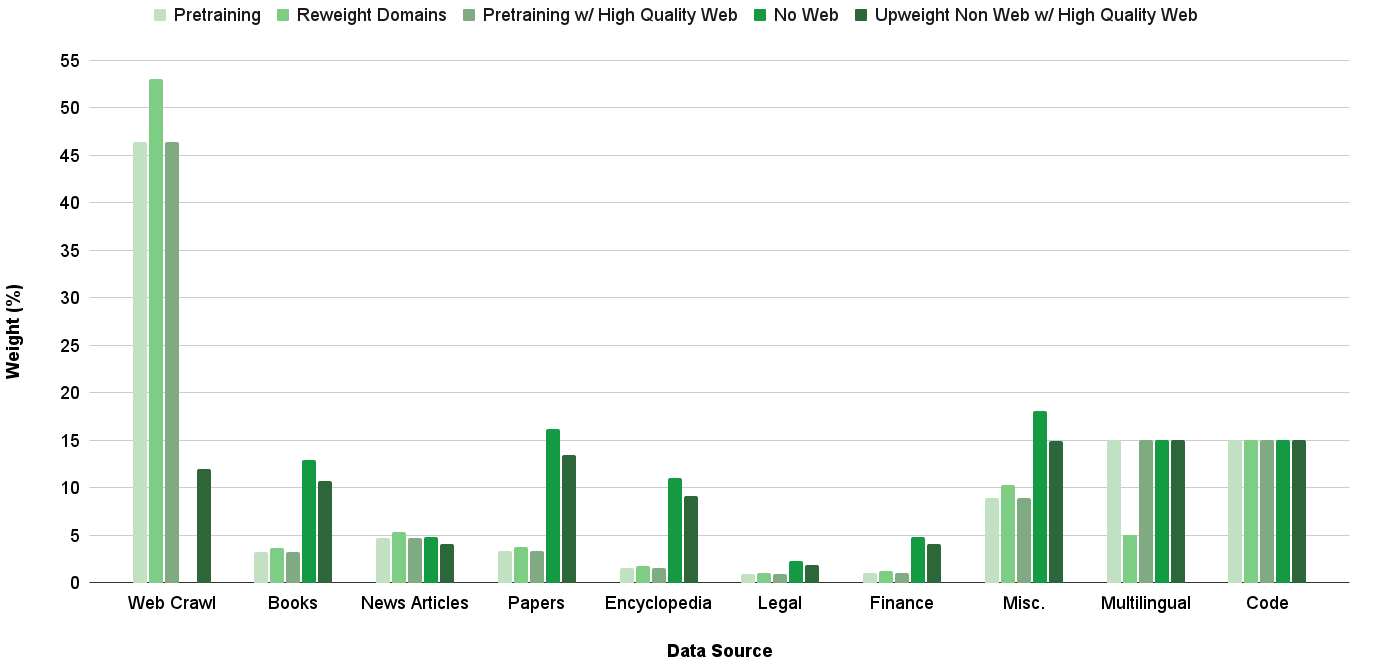}
    \caption{Breakdown of the various distributions considered for the General Blend (GB). We use Upweight Non Web w/ High Quality Web as the GB moving forward given its strong performance across all evaluation areas.}
    \label{fig:gb_distrs}
\end{figure*}

A crucial component of any training run is the data distribution -- it defines the information which a model sees and directly impacts the model's capabilities. As continuous pretraining builds on top of a model which has already seen a given pretraining distribution, it is important to define a data distribution which allows the model to learn new concepts without also deviating too far from the pretraining distribution such that the model begins to experience training instability and accuracy regression. Through a series of runs which tackle what compositions of data distributions best improve the abilities of a pretrained model, we identify general characteristics that can be applied across most continuous pretraining scenarios. In these experiments, we use a learning rate schedule that starts from $\eta_{min}$ and decays to 0 with cosine annealing.

First, we examine if the inclusion of QA data, which improves the ability of a model to extract stored knowledge \citep{allenzhu2023physics}, improves model accuracy. Coupled with this question is another on how to best incorporate the QA data, or more generally any dataset which is not contained within the pretraining data distribution, into the continued training run: immediately at the beginning and throughout the entirety of continued training, or rather reserved till the end of continued training following a curriculum learning setup \citep{soviany2022curriculum, blakeney2024does}. We hypothesize that inclusion of new data sources at the beginning of continued pretraining allows for the model to best learn the new information, but may cause learning instabilities that could be mitigated by showing the new dataset at the end of the run when the learning rate is less aggressive. To answer these questions, we compare continued training entirely with the pretraining data blend, entirely with a QA data blend, and with a mix of the pretraining and QA data blends where we start with the pretraining blend and switch to the QA data blend late in the training run. The QA data blend in this scenario adds the QA dataset to the pretraining data distribution with a weight of 10\%.

\begin{table}[!h]
\centering
  \begin{tabular}{lr}
   \toprule
    \textbf{Data Blend} & \textbf{Avg. Acc.} \\
    \midrule
    Pretraining &  51.5  \\
    QA &   53.4  \\
    Pretraining (250B), QA (50B) &  \textbf{54.3}  \\
    \bottomrule
  \end{tabular}
  \caption{Using two data distributions, with the QA data appearing in the latter, leads to the largest improvement via continued pretraining. () indicates the number of training tokens for each blend. Per-task evaluations scores are shared in Table \ref{tab:data_distr_exps_results_detailed}. }
  \label{tab:initial_exps_results}
\end{table}

Table \ref{tab:initial_exps_results} illustrates that the incorporation of QA data markedly outperforms solely using existing data from the pretraining set. Additionally, first using the pretraining data blend for the majority of training tokens before transitioning to the QA data blend at the end of continued pretraining exhibits improved accuracy compared to using the QA blend throughout the entirety of training. This indicates that continued pretraining runs should begin with a data distribution which more closely aligns to the pretraining one followed by a blend that then introduces new data. Moving forward, we refer to the initial blend as the general blend, GB, and the latter blend as the QA blend, QB, and discuss how they can be refined to realize further improvements.

We hypothesize that the optimal GB will be one which places greater emphasis on high quality data sources and areas of model weakness, without deviating too far from the pretraining distribution. Such a blend will enhance knowledge in needed areas and prime the model for the QB blend without worry of experiencing large training instabilities. Figure \ref{fig:gb_distrs} illustrates the various GB distributions we consider; in addition to upweighting sources of interest, we either subset web crawl to just high quality documents, as identified by being in the bottom quartile of perplexity scores from a KenLM model \citep{heafield2011kenlm} trained on Wikipedia, or remove web crawl altogether. Experimenting with the various GB distributions for all 300B tokens of continued training, Table \ref{tab:general_blend_results} shows that
each improves upon the pretraining distribution. Even though it does not achieve the highest average accuracy, we choose Upweight Non Web with High Quality Web as the GB moving forward, because compared to others, it most consistently achieves high scores across all considered tasks as shown in Table \ref{tab:data_distr_exps_results_detailed}.

\begin{table}[!h]
\centering
  \begin{tabular}{lr}
   \toprule
    \textbf{Data Blend} & \textbf{Avg. Acc.} \\
    \midrule
    Pretraining &  51.5  \\
    Reweight Domains &   51.7 \\
    Pretraining w/ High Quality Web  &  52.5  \\
    No Web  &  \textbf{52.9}  \\
    UW Non Web w/ High Quality Web  &  52.0  \\
    \bottomrule
  \end{tabular}
  \caption{Evaluation results of various GB candidate distributions. Per-task evaluations scores are shared in Table \ref{tab:data_distr_exps_results_detailed} }
  \label{tab:general_blend_results}
\end{table}

\begin{figure}[t]
    \centering
    \includegraphics[width=\linewidth]{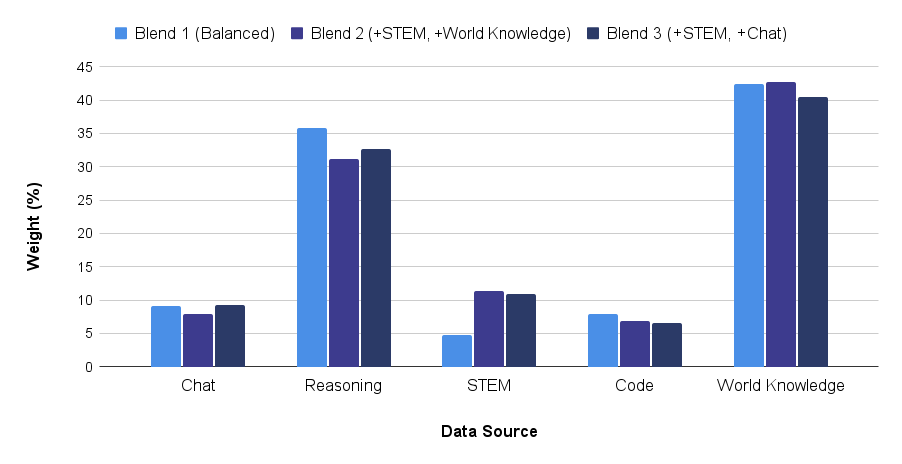}
    \caption{Various distributions of QA data. We use Blend 3.}
    \label{fig:qb_qa_distr}
\end{figure}

With a GB distribution in place, we now look to define the QB distribution by first refining the weights placed on the sources within the QA data and then optimizing the QB distribution as a whole. In the initial QB distribution, the QA data was added as is, and this weighting is shown as QA blend 1 in Figure \ref{fig:qb_qa_distr}. Given that the pretrained model struggles on STEM tasks, we create two additional blends that both upweight the QA STEM data while either maintaining the original weight of QA world knowledge, blend 2, or QA chat, blend 3, data as seen in Figure \ref{fig:qb_qa_distr}. We choose to maintain the weight in world knowledge and chat information as such examples cover a broad range of topics and help better align model responses to questions respectively. Table \ref{tab:qa_blends_results} highlights that upon adding each of the QA blends to the initial QB distribution following 250B tokens of the identified GB, QA data that emphasizes both STEM and chat information leads to the best results.


\begin{table}[!h]
\centering
  \begin{tabular}{lr}
   \toprule
    \textbf{Data Blend} & \textbf{Avg. Acc.} \\
    \midrule
    QA 1 &  54.3  \\
    QA 2 (+STEM, +World Knowledge)  &   53.0 \\
    QA 3 (+STEM, +Chat)  &  \textbf{54.9}  \\
    \bottomrule
  \end{tabular}
  \caption{Evaluation results of various QA blend candidates. Per-task evaluations scores are shared in Table \ref{tab:data_distr_exps_results_detailed} }
  \label{tab:qa_blends_results}
\end{table}

\begin{figure*}[t]
    \centering
    \includegraphics[width=\linewidth]{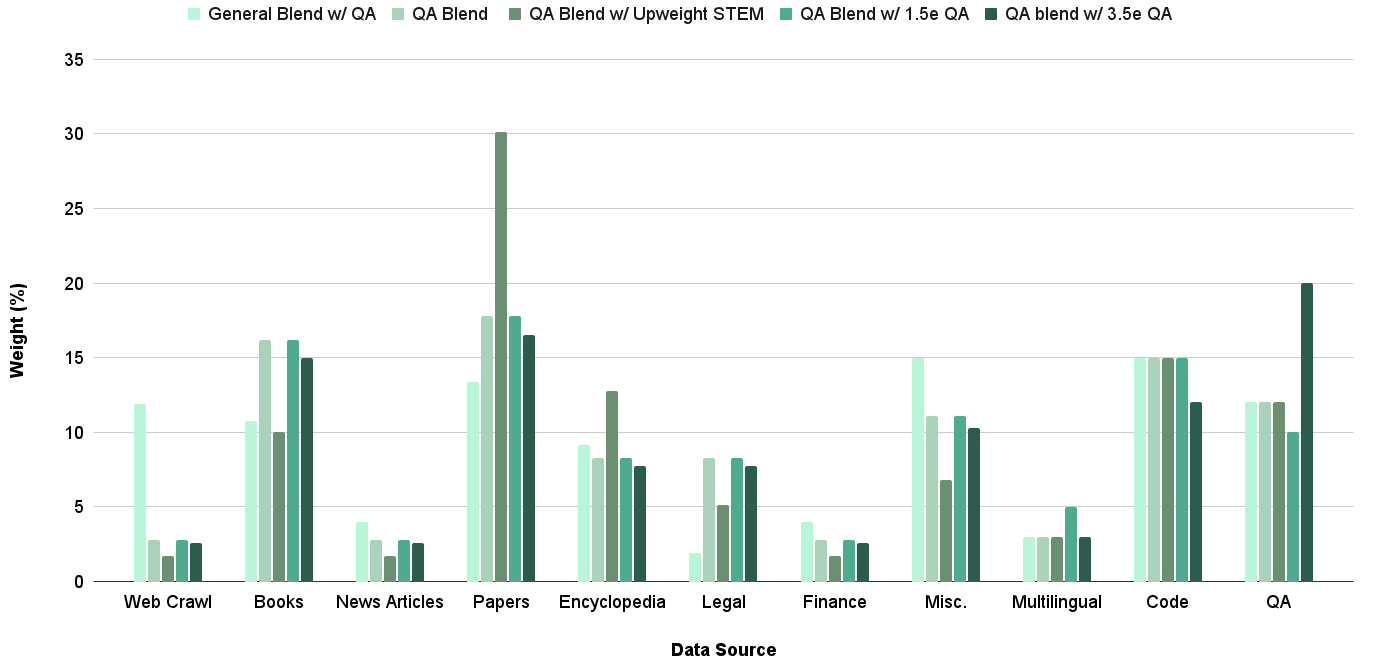}
    \caption{Breakdown of the various distributions considered for the QB. $N$e refers to $N$ epochs of the QA data. The final chosen distribution is shown as QA Blend which used 2 epochs of QA data.}
    \label{fig:qb_distrs}
\end{figure*}

We now incorporate the QA data within the overall QB distribution. In previous runs, the QB distribution, aside from the QA dataset, exactly mirrored the pretraining set. We define a new series of distributions based on more aggressive upweighting of sources in areas of model weakness and amount of weight placed on the QA dataset as seen in Figure \ref{fig:qb_distrs}. Table \ref{tab:qa_blends_full_results}
details that the aggressive weighting in the QB is beneficial, and we use the QB termed QA blend moving forward. With refined GB and QB distributions, the average evaluation accuracy has improved from 48.9 for the pretrained model to 55.4, a 13\% improvement.
\begin{table}[!h]
\centering
  \begin{tabular}{lr}
   \toprule
    \textbf{Data Blend} & \textbf{Avg. Acc.} \\
    \midrule
    Pretraining blend w/ QA data & 54.3   \\
    General blend w/ QA data & 54.2   \\
    QA  & \textbf{55.4}  \\
    QA w/ Upweighted STEM   &  54.4  \\
    QA w/ 1.5e QA data   &  54.9  \\
    QA w/ 3.5e QA data   & 54.4  \\
    \bottomrule
  \end{tabular}
  \caption{Evaluation results of various QB candidate distributions. Per-task evaluations scores are shared in Table \ref{tab:data_distr_exps_results_detailed} }
  \label{tab:qa_blends_full_results}
\end{table}

\begin{figure}[t]
    \centering
    \includegraphics[width=\linewidth]{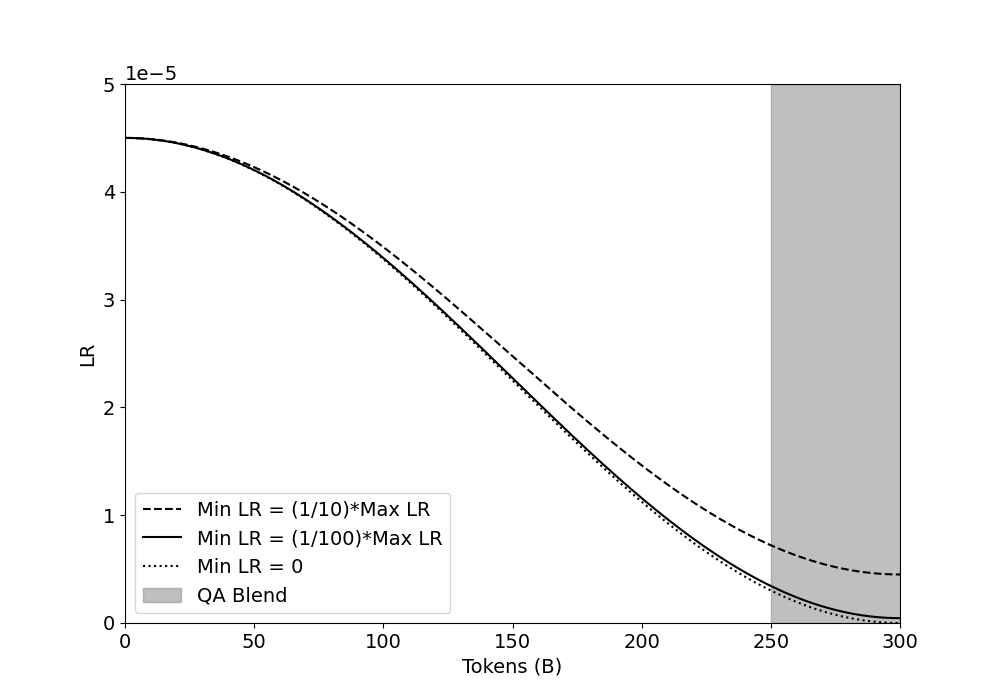}
    \caption{Cosine decay schedules with a Max LR of $4.5e\text{-}5$. Each schedule differently prioritizes LR magnitude and slope of decay.}
    \label{fig:decay_lrs}
\end{figure}

\subsection{Learning Rate Schedule}

The learning rate schedule greatly impacts the training dynamics and efficacy of continued pretraining \citep{gupta2023continual, ibrahim2024simple, winata2023overcoming}.

In our above continued pretraining experiments, the learning rate schedule
starts at a maximum LR of $\eta_{max_{\text{ct}}} = 4.5e\text{-}5$, which is equal to $\eta_{min}$, and decays to a minimum LR of 0 using cosine annealing. As seen in Figure \ref{fig:decay_lrs}, a minimum LR of 0 facilitates a steep slope of decay but the magnitude of LR is severely impacted, especially over the tokens where the QB is used which may impact the model's ability to extract full utility from the QA data. To understand the trade-off between these two characteristics of the learning rate schedule in continued pretraining runs, we experiment with two additional minimum learning rate values: $\frac{\eta_{max_{\text{ct}}}}{10} = 4.5e\text{-}6$ and $\frac{\eta_{max_{\text{ct}}}}{100} = 4.5e\text{-}7$. 

\begin{table}[!h]
\centering
  \begin{tabular}{lr}
   \toprule
    \textbf{LR Schedule} & \textbf{Avg. Acc.} \\
    \midrule
    Decay to $\frac{\eta_{max_{\text{ct}}}}{10}$ &  54.8  \\
    Decay to $\frac{\eta_{max_{\text{ct}}}}{100}$ &   \textbf{55.7} \\
    Decay to 0  &  55.4  \\
    \bottomrule
  \end{tabular}
  \caption{Evaluation results of learning rate schedules with varying Min LR values. Per-task evaluations scores are shared in Table \ref{tab:lr_decay_results_detailed} }
  \label{tab:lr_decay_results}
\end{table}

Table \ref{tab:lr_decay_results} highlights that it is in fact best to strike a middle ground between magnitude of LR and slope of decay, as a minimum LR of $\frac{\eta_{max_{\text{ct}}}}{100}$ achieves the best accuracy. Such a minimum LR value allows for a learning rate schedule that has reasonable decay over the QB tokens, unlike when using a minimum LR of $\frac{\eta_{max_{\text{ct}}}}{10}$, without severely sacrificing on magnitude of LR, as was the case with a minimum LR of 0.

Experiments with varying learning rate warmup and maximum LR values led to accuracy regressions compared to the schedule detailed above.
In addition, we ran ablations with a different annealing schedule, WSD \citep{hu2024minicpm}, however the results were not competitive to cosine annealing. Full details and results for both studies are shared in Appendix \ref{sec:appendix_experiments_learning_rates}.



\subsection{Switch of Data Distributions}
\label{sec:experiments_distribution_switch}

\begin{figure}[h]
    \centering
    \includegraphics[width=\linewidth]{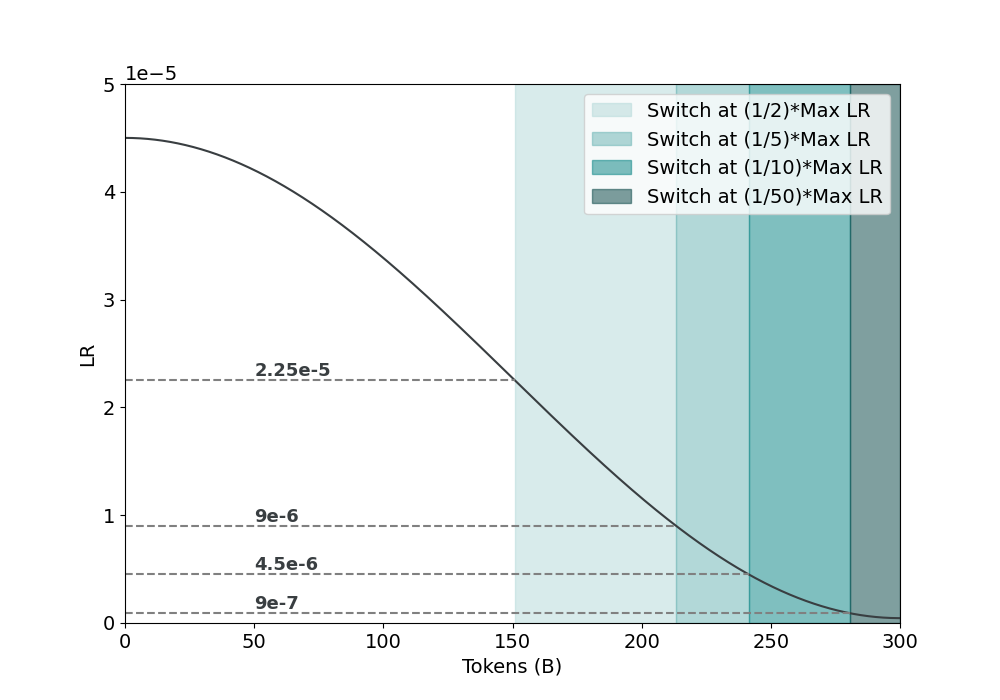}
    \caption{How the number of QB tokens, the shaded region, varies based on different distribution switch points.}
    \label{fig:lr_distribution_shift}
\end{figure}

Until this point, we have been switching between the GB and the QB after 250B tokens of continued pretraining. We believe this to be sub-optimal, as it is unclear how switching between distributions after a fixed number of tokens can be easily translated to continued training runs of different token horizons. We hypothesize that the optimal point for switching between the data distributions depends upon the learning rate schedule. Figure \ref{fig:lr_distribution_shift} highlights how both the number of tokens and learning rate values for the QB blend would differ if the distribution switch occurred at progressively smaller fractions of the maximum LR. As the fraction goes to 0, both the slope of decay and magnitude of the learning rate shrink, meaning that there likely is an optimal point in the learning rate curve where both of these characteristics are still conducive to enable learning but also not too aggressive to the point where the data shift in the QB distribution causes training instability.

Table \ref{tab:distribution_switch} highlights that switching between the GB and QB at $\frac{\eta_{max_{\text{ct}}}}{5}$ achieves the best accuracy and improves upon the heuristically chosen switch point by 0.4 points on average. Wanting to confirm this distribution switch point holds at differing amounts of continued pretraining tokens, we ran an ablation on a scale of 100B tokens and found that $\frac{\eta_{max_{\text{ct}}}}{5}$ again maximized the results as seen in Table \ref{tab:distribution_switch_100B}. 
\begin{table}[!h]
\centering
  \begin{tabular}{lr}
   \toprule
    \textbf{Distribution Switch} & \textbf{Avg. Acc.} \\
    \midrule
    At $\eta_{max_{\text{ct}}}$ (from step 0) &  52.8  \\
    At $\frac{\eta_{max_{\text{ct}}}}{2}$  &   54.7 \\
    At $\frac{\eta_{max_{\text{ct}}}}{5}$  &   \textbf{56.1} \\
    At $\frac{\eta_{max_{\text{ct}}}}{10}$  &   55.0 \\
    At $\frac{\eta_{max_{\text{ct}}}}{50}$  &   54.6 \\
    \bottomrule
  \end{tabular}
  \caption{Evaluation results of varying distribution switch points. Per-task evaluations scores are shared in Table \ref{tab:distribution_switch_results_detailed} }
  \label{tab:distribution_switch}
\end{table}

This finalizes our continued pretraining recipe. We highlight the utility of this recipe as it allows the model to achieve an average accuracy of 56.1, which improves upon the natural baseline of continued training on the pretraining distribution, as shared in Table \ref{tab:initial_exps_results}, by 9\%. 

\section{Ablations}

\subsection{Varying Token Horizons}

 We show the efficacy of the identified continued pretraining recipe when used at varying numbers of continued training tokens. Table  \ref{ct_varying_tokens_results} illustrates that on continued training horizons from 100B to 1T tokens, the identified recipe consistently achieves improved evaluation results -- realizing a 16\% gain over the pretrained model when using 1T tokens of continued training. We do note that the slope in accuracy improvement from 300B to 1T tokens is lower than that from 100B to 300B tokens, we hypothesize that as we are mainly reusing documents from the pretraining set when doing a large number of continued training tokens the repeated number of epochs on the same data sources have decreasing marginal utility. 


\begin{table}[!h]
\centering
  \begin{tabular}{lcc}
   \toprule
    \textbf{Num CPT Tokens} & \textbf{MMLU} & \textbf{Avg. Acc.} \\
    \midrule
    0B & 59.3 & 48.9 \\
    100B &  63.0  &  55.0  \\
    300B  &   63.8 &  56.1 \\
    1T  &   \textbf{65.3} &  \textbf{56.8} \\
    \bottomrule
  \end{tabular}
  \caption{Performance of the continuous pretraining (CPT) recipe across different token horizons. Per-task evaluations scores are shared in Table \ref{tab:ct_varying_tokens_results_detailed} }
  \label{ct_varying_tokens_results}
\end{table}

\subsection{Document Mining}


In an effort to improve the utility of the data sources that are seen for multiple epochs in long horizon continued pretraining runs, we aim to find a subset of examples that are most helpful for model improvement. As the QA dataset was shown to significantly boost model accuracies, we hypothesize that restricting each pretraining data source to the set of documents which are most similar to the QA examples would be beneficial. To do so, we use the E5-large-v2 \citep{wang2022text} text embedding model to obtain an embedding for each document in our pretraining and QA sets. Using the Faiss library \citep{johnson2017billionscale}, we efficiently perform a 50-nearest neighbor search across all these embeddings to obtain the 50 most similar, non-QA documents to each example in the QA set. The identified subset of examples constitutes 60B tokens, and we term this approach document mining.

Table \ref{ct_mined_documents_results} shows a training run where we replace all non-QA data sources in the QB distribution solely with the examples identified via document mining. We find that these documents substantially improve the performance of the continued pretraining run and believe that document mining is a viable approach at extracting further utility from existing data sources. 


\begin{table}[!h]
\centering
  \begin{tabular}{lcc}
   \toprule
    \textbf{Blend} & \textbf{MMLU} & \textbf{Avg. Acc.} \\
    \midrule
    CT 1T & 65.3 & 56.8 \\
    CT 1T w/ Mined Docs &  \textbf{66.6}  &  \textbf{57.9}  \\
    \bottomrule
  \end{tabular}
  \caption{Mining examples related to QA documents further improves accuracy. Per-task evaluations scores are shared in  Table \ref{tab:ct_mined_documents_results_detailed} }
  \label{ct_mined_documents_results}
\end{table}

\section{Conclusion}

We investigate how to effectively continue training LMs to improve upon their existing capabilities. Our experiments show that it is especially important to carefully define the data distribution and learning rate decay schedule used during continued pretraining so that the model is able to smoothly transition away from the pretraining distribution and better learn the newly emphasized data sources.  
With these findings we propose a general recipe that model developers can use in order to perform continued pretraining on top of their own LMs and show that for our base model, we are able to improve cumulative accuracy by over 18\%. We hope that this will be a starting point to enable future LMs to be developed through the reuse of existing models rather than retraining from scratch.

\section*{Limitations}

In the development of our continued pretraining recipe, we only experiment along the axes of data distributions and hyperparameter configurations. Although we did not include them within our study, there may be added benefit in exploring other aspects such as altering the learning algorithm. Additionally, given that our study is conducted on top of a model with a given configuration and which was pretrained using a certain data distribution, the results that we highlight are likely to not extrapolate well when used in settings highly divergent from the one utilized in the study. Finally, we limited our goal within continued pretraining to improving the general purpose capabilities of the pretrained model; however, there are many additional angles when considering model reuse such as domain specialization and the efficient addition of new knowledge into existing models. 

\clearpage
\newpage
\bibliography{custom}

\begin{thebibliography}{57}
\providecommand{\natexlab}[1]{#1}

\bibitem[{Ainslie et~al.(2023)Ainslie, Lee-Thorp, de~Jong, Zemlyanskiy, Lebr{\'o}n, and Sanghai}]{ainslie2023gqa}
Joshua Ainslie, James Lee-Thorp, Michiel de~Jong, Yury Zemlyanskiy, Federico Lebr{\'o}n, and Sumit Sanghai. 2023.
\newblock {GQA: Training Generalized Multi-Query Transformer Models from Multi-Head Checkpoints}.
\newblock \emph{arXiv preprint arXiv:2305.13245}.

\bibitem[{Allen-Zhu and Li(2023)}]{allenzhu2023physics}
Zeyuan Allen-Zhu and Yuanzhi Li. 2023.
\newblock \href {https://arxiv.org/abs/2309.14316} {Physics of language models: Part 3.1, knowledge storage and extraction}.
\newblock \emph{Preprint}, arXiv:2309.14316.

\bibitem[{Anthropic(2024)}]{claude32024}
Anthropic. 2024.
\newblock {The Claude 3 Model Family: Opus, Sonnet, Haiku}.

\bibitem[{Blakeney et~al.(2024)Blakeney, Paul, Larsen, Owen, and Frankle}]{blakeney2024does}
Cody Blakeney, Mansheej Paul, Brett~W. Larsen, Sean Owen, and Jonathan Frankle. 2024.
\newblock \href {https://arxiv.org/abs/2406.03476} {Does your data spark joy? performance gains from domain upsampling at the end of training}.
\newblock \emph{Preprint}, arXiv:2406.03476.

\bibitem[{Brown et~al.(2020)Brown, Mann, Ryder, Subbiah, Kaplan, Dhariwal, Neelakantan, Shyam, Sastry, Askell, Agarwal, Herbert-Voss, Krueger, Henighan, Child, Ramesh, Ziegler, Wu, Winter, Hesse, Chen, Sigler, Litwin, Gray, Chess, Clark, Berner, McCandlish, Radford, Sutskever, and Amodei}]{brown2020language}
Tom~B. Brown, Benjamin Mann, Nick Ryder, Melanie Subbiah, Jared Kaplan, Prafulla Dhariwal, Arvind Neelakantan, Pranav Shyam, Girish Sastry, Amanda Askell, Sandhini Agarwal, Ariel Herbert-Voss, Gretchen Krueger, Tom Henighan, Rewon Child, Aditya Ramesh, Daniel~M. Ziegler, Jeffrey Wu, Clemens Winter, Christopher Hesse, Mark Chen, Eric Sigler, Mateusz Litwin, Scott Gray, Benjamin Chess, Jack Clark, Christopher Berner, Sam McCandlish, Alec Radford, Ilya Sutskever, and Dario Amodei. 2020.
\newblock \href {https://arxiv.org/abs/2005.14165} {Language models are few-shot learners}.
\newblock \emph{Preprint}, arXiv:2005.14165.

\bibitem[{Caccia et~al.(2021)Caccia, Rodriguez, Ostapenko, Normandin, Lin, Caccia, Laradji, Rish, Lacoste, Vazquez, and Charlin}]{caccia2021online}
Massimo Caccia, Pau Rodriguez, Oleksiy Ostapenko, Fabrice Normandin, Min Lin, Lucas Caccia, Issam Laradji, Irina Rish, Alexandre Lacoste, David Vazquez, and Laurent Charlin. 2021.
\newblock \href {https://arxiv.org/abs/2003.05856} {Online fast adaptation and knowledge accumulation: a new approach to continual learning}.
\newblock \emph{Preprint}, arXiv:2003.05856.

\bibitem[{Chaudhry et~al.(2019)Chaudhry, Rohrbach, Elhoseiny, Ajanthan, Dokania, Torr, and Ranzato}]{chaudhry2019tiny}
Arslan Chaudhry, Marcus Rohrbach, Mohamed Elhoseiny, Thalaiyasingam Ajanthan, Puneet~K. Dokania, Philip H.~S. Torr, and Marc'Aurelio Ranzato. 2019.
\newblock \href {https://arxiv.org/abs/1902.10486} {On tiny episodic memories in continual learning}.
\newblock \emph{Preprint}, arXiv:1902.10486.

\bibitem[{Chen et~al.(2021)Chen, Tworek, Jun, Yuan, de~Oliveira~Pinto, Kaplan, Edwards, Burda, Joseph, Brockman, Ray, Puri, Krueger, Petrov, Khlaaf, Sastry, Mishkin, Chan, Gray, Ryder, Pavlov, Power, Kaiser, Bavarian, Winter, Tillet, Such, Cummings, Plappert, Chantzis, Barnes, Herbert-Voss, Guss, Nichol, Paino, Tezak, Tang, Babuschkin, Balaji, Jain, Saunders, Hesse, Carr, Leike, Achiam, Misra, Morikawa, Radford, Knight, Brundage, Murati, Mayer, Welinder, McGrew, Amodei, McCandlish, Sutskever, and Zaremba}]{chen2021evaluating}
Mark Chen, Jerry Tworek, Heewoo Jun, Qiming Yuan, Henrique~Ponde de~Oliveira~Pinto, Jared Kaplan, Harri Edwards, Yuri Burda, Nicholas Joseph, Greg Brockman, Alex Ray, Raul Puri, Gretchen Krueger, Michael Petrov, Heidy Khlaaf, Girish Sastry, Pamela Mishkin, Brooke Chan, Scott Gray, Nick Ryder, Mikhail Pavlov, Alethea Power, Lukasz Kaiser, Mohammad Bavarian, Clemens Winter, Philippe Tillet, Felipe~Petroski Such, Dave Cummings, Matthias Plappert, Fotios Chantzis, Elizabeth Barnes, Ariel Herbert-Voss, William~Hebgen Guss, Alex Nichol, Alex Paino, Nikolas Tezak, Jie Tang, Igor Babuschkin, Suchir Balaji, Shantanu Jain, William Saunders, Christopher Hesse, Andrew~N. Carr, Jan Leike, Josh Achiam, Vedant Misra, Evan Morikawa, Alec Radford, Matthew Knight, Miles Brundage, Mira Murati, Katie Mayer, Peter Welinder, Bob McGrew, Dario Amodei, Sam McCandlish, Ilya Sutskever, and Wojciech Zaremba. 2021.
\newblock \href {https://arxiv.org/abs/2107.03374} {Evaluating large language models trained on code}.
\newblock \emph{Preprint}, arXiv:2107.03374.

\bibitem[{Chowdhery et~al.(2022)Chowdhery, Narang, Devlin, Bosma, Mishra, Roberts, Barham, Chung, Sutton, Gehrmann et~al.}]{chowdhery2022palm}
Aakanksha Chowdhery, Sharan Narang, Jacob Devlin, Maarten Bosma, Gaurav Mishra, Adam Roberts, Paul Barham, Hyung~Won Chung, Charles Sutton, Sebastian Gehrmann, et~al. 2022.
\newblock {PaLM: Scaling Language Modeling with Pathways}.
\newblock \emph{arXiv preprint arXiv:2204.02311}.

\bibitem[{DeepSeek-AI et~al.(2024)DeepSeek-AI, :, Bi, Chen, Chen, Chen, Dai, Deng, Ding, Dong, Du, Fu, Gao, Gao, Gao, Ge, Guan, Guo, Guo, Hao, Hao, He, Hu, Huang, Li, Li, Li, Li, Li, Liang, Lin, Liu, Liu, Liu, Liu, Liu, Liu, Lu, Lu, Luo, Ma, Nie, Pei, Piao, Qiu, Qu, Ren, Ren, Ruan, Sha, Shao, Song, Su, Sun, Sun, Tang, Wang, Wang, Wang, Wang, Wang, Wu, Wu, Xie, Xie, Xie, Xiong, Xu, Xu, Xu, Yang, You, Yu, Yu, Zhang, Zhang, Zhang, Zhang, Zhang, Zhang, Zhang, Zhang, Zhao, Zhao, Zhou, Zhou, Zhu, and Zou}]{deepseekai2024deepseek}
DeepSeek-AI, :, Xiao Bi, Deli Chen, Guanting Chen, Shanhuang Chen, Damai Dai, Chengqi Deng, Honghui Ding, Kai Dong, Qiushi Du, Zhe Fu, Huazuo Gao, Kaige Gao, Wenjun Gao, Ruiqi Ge, Kang Guan, Daya Guo, Jianzhong Guo, Guangbo Hao, Zhewen Hao, Ying He, Wenjie Hu, Panpan Huang, Erhang Li, Guowei Li, Jiashi Li, Yao Li, Y.~K. Li, Wenfeng Liang, Fangyun Lin, A.~X. Liu, Bo~Liu, Wen Liu, Xiaodong Liu, Xin Liu, Yiyuan Liu, Haoyu Lu, Shanghao Lu, Fuli Luo, Shirong Ma, Xiaotao Nie, Tian Pei, Yishi Piao, Junjie Qiu, Hui Qu, Tongzheng Ren, Zehui Ren, Chong Ruan, Zhangli Sha, Zhihong Shao, Junxiao Song, Xuecheng Su, Jingxiang Sun, Yaofeng Sun, Minghui Tang, Bingxuan Wang, Peiyi Wang, Shiyu Wang, Yaohui Wang, Yongji Wang, Tong Wu, Y.~Wu, Xin Xie, Zhenda Xie, Ziwei Xie, Yiliang Xiong, Hanwei Xu, R.~X. Xu, Yanhong Xu, Dejian Yang, Yuxiang You, Shuiping Yu, Xingkai Yu, B.~Zhang, Haowei Zhang, Lecong Zhang, Liyue Zhang, Mingchuan Zhang, Minghua Zhang, Wentao Zhang, Yichao Zhang, Chenggang Zhao, Yao Zhao, Shangyan Zhou, Shunfeng
  Zhou, Qihao Zhu, and Yuheng Zou. 2024.
\newblock \href {https://arxiv.org/abs/2401.02954} {Deepseek llm: Scaling open-source language models with longtermism}.
\newblock \emph{Preprint}, arXiv:2401.02954.

\bibitem[{El-Kishky et~al.(2019)El-Kishky, Chaudhary, Guzm{\'a}n, and Koehn}]{el2019ccaligned}
Ahmed El-Kishky, Vishrav Chaudhary, Francisco Guzm{\'a}n, and Philipp Koehn. 2019.
\newblock Ccaligned: A massive collection of cross-lingual web-document pairs.
\newblock \emph{arXiv preprint arXiv:1911.06154}.

\bibitem[{French(1999)}]{FRENCH1999128}
Robert~M. French. 1999.
\newblock \href {https://doi.org/10.1016/S1364-6613(99)01294-2} {Catastrophic forgetting in connectionist networks}.
\newblock \emph{Trends in Cognitive Sciences}, 3(4):128--135.

\bibitem[{Gao et~al.(2020)Gao, Biderman, Black, Golding, Hoppe, Foster, Phang, He, Thite, Nabeshima, Presser, and Leahy}]{pile-dataset-2020}
Leo Gao, Stella Biderman, Sid Black, Laurence Golding, Travis Hoppe, Charles Foster, Jason Phang, Horace He, Anish Thite, Noa Nabeshima, Shawn Presser, and Connor Leahy. 2020.
\newblock The {P}ile: An 800gb dataset of diverse text for language modeling.
\newblock \emph{arXiv preprint arXiv:2101.00027}.

\bibitem[{Gemma~Team(2024)}]{gemma24}
Google~DeepMind Gemma~Team. 2024.
\newblock {Gemma: Open Models Based on Gemini Research and Technology}.

\bibitem[{Gupta et~al.(2023)Gupta, Thérien, Ibrahim, Richter, Anthony, Belilovsky, Rish, and Lesort}]{gupta2023continual}
Kshitij Gupta, Benjamin Thérien, Adam Ibrahim, Mats~L. Richter, Quentin Anthony, Eugene Belilovsky, Irina Rish, and Timothée Lesort. 2023.
\newblock \href {https://arxiv.org/abs/2308.04014} {Continual pre-training of large language models: How to (re)warm your model?}
\newblock \emph{Preprint}, arXiv:2308.04014.

\bibitem[{Gururangan et~al.(2020)Gururangan, Marasovi{\'c}, Swayamdipta, Lo, Beltagy, Downey, and Smith}]{gururangan-etal-2020-dont}
Suchin Gururangan, Ana Marasovi{\'c}, Swabha Swayamdipta, Kyle Lo, Iz~Beltagy, Doug Downey, and Noah~A. Smith. 2020.
\newblock \href {https://doi.org/10.18653/v1/2020.acl-main.740} {Don{'}t stop pretraining: Adapt language models to domains and tasks}.
\newblock In \emph{Proceedings of the 58th Annual Meeting of the Association for Computational Linguistics}, pages 8342--8360, Online. Association for Computational Linguistics.

\bibitem[{Heafield(2011)}]{heafield2011kenlm}
Kenneth Heafield. 2011.
\newblock Kenlm: Faster and smaller language model queries.
\newblock In \emph{Proceedings of the sixth workshop on statistical machine translation}, pages 187--197.

\bibitem[{Hendrycks et~al.(2020)Hendrycks, Burns, Basart, Zou, Mazeika, Song, and Steinhardt}]{hendrycks2020measuring}
Dan Hendrycks, Collin Burns, Steven Basart, Andy Zou, Mantas Mazeika, Dawn Song, and Jacob Steinhardt. 2020.
\newblock {Measuring Massive Multitask Language Understanding}.
\newblock \emph{arXiv preprint arXiv:2009.03300}.

\bibitem[{Hinton et~al.(2015)Hinton, Vinyals, and Dean}]{hinton2015distilling}
Geoffrey Hinton, Oriol Vinyals, and Jeff Dean. 2015.
\newblock \href {https://arxiv.org/abs/1503.02531} {Distilling the knowledge in a neural network}.
\newblock \emph{Preprint}, arXiv:1503.02531.

\bibitem[{Hu et~al.(2024)Hu, Tu, Han, He, Cui, Long, Zheng, Fang, Huang, Zhao, Zhang, Thai, Zhang, Wang, Yao, Zhao, Zhou, Cai, Zhai, Ding, Jia, Zeng, Li, Liu, and Sun}]{hu2024minicpm}
Shengding Hu, Yuge Tu, Xu~Han, Chaoqun He, Ganqu Cui, Xiang Long, Zhi Zheng, Yewei Fang, Yuxiang Huang, Weilin Zhao, Xinrong Zhang, Zheng~Leng Thai, Kaihuo Zhang, Chongyi Wang, Yuan Yao, Chenyang Zhao, Jie Zhou, Jie Cai, Zhongwu Zhai, Ning Ding, Chao Jia, Guoyang Zeng, Dahai Li, Zhiyuan Liu, and Maosong Sun. 2024.
\newblock \href {https://arxiv.org/abs/2404.06395} {Minicpm: Unveiling the potential of small language models with scalable training strategies}.
\newblock \emph{Preprint}, arXiv:2404.06395.

\bibitem[{Ibrahim et~al.(2024)Ibrahim, Thérien, Gupta, Richter, Anthony, Lesort, Belilovsky, and Rish}]{ibrahim2024simple}
Adam Ibrahim, Benjamin Thérien, Kshitij Gupta, Mats~L. Richter, Quentin Anthony, Timothée Lesort, Eugene Belilovsky, and Irina Rish. 2024.
\newblock \href {https://arxiv.org/abs/2403.08763} {Simple and scalable strategies to continually pre-train large language models}.
\newblock \emph{Preprint}, arXiv:2403.08763.

\bibitem[{Jang et~al.(2023)Jang, Ye, Lee, Yang, Shin, Han, Kim, and Seo}]{jang2023temporalwiki}
Joel Jang, Seonghyeon Ye, Changho Lee, Sohee Yang, Joongbo Shin, Janghoon Han, Gyeonghun Kim, and Minjoon Seo. 2023.
\newblock \href {https://arxiv.org/abs/2204.14211} {Temporalwiki: A lifelong benchmark for training and evaluating ever-evolving language models}.
\newblock \emph{Preprint}, arXiv:2204.14211.

\bibitem[{Jang et~al.(2022)Jang, Ye, Yang, Shin, Han, Kim, Choi, and Seo}]{jang2022continual}
Joel Jang, Seonghyeon Ye, Sohee Yang, Joongbo Shin, Janghoon Han, Gyeonghun Kim, Stanley~Jungkyu Choi, and Minjoon Seo. 2022.
\newblock \href {https://arxiv.org/abs/2110.03215} {Towards continual knowledge learning of language models}.
\newblock \emph{Preprint}, arXiv:2110.03215.

\bibitem[{Jin et~al.(2022)Jin, Zhang, Zhu, Xiao, Li, Wei, Arnold, and Ren}]{jin2022lifelong}
Xisen Jin, Dejiao Zhang, Henghui Zhu, Wei Xiao, Shang-Wen Li, Xiaokai Wei, Andrew Arnold, and Xiang Ren. 2022.
\newblock \href {https://arxiv.org/abs/2110.08534} {Lifelong pretraining: Continually adapting language models to emerging corpora}.
\newblock \emph{Preprint}, arXiv:2110.08534.

\bibitem[{Johnson et~al.(2017)Johnson, Douze, and Jégou}]{johnson2017billionscale}
Jeff Johnson, Matthijs Douze, and Hervé Jégou. 2017.
\newblock \href {https://arxiv.org/abs/1702.08734} {Billion-scale similarity search with gpus}.
\newblock \emph{Preprint}, arXiv:1702.08734.

\bibitem[{Ke et~al.(2023)Ke, Shao, Lin, Konishi, Kim, and Liu}]{ke2023continual}
Zixuan Ke, Yijia Shao, Haowei Lin, Tatsuya Konishi, Gyuhak Kim, and Bing Liu. 2023.
\newblock \href {https://arxiv.org/abs/2302.03241} {Continual pre-training of language models}.
\newblock \emph{Preprint}, arXiv:2302.03241.

\bibitem[{Kudo and Richardson(2018)}]{kudo2018sentencepiece}
Taku Kudo and John Richardson. 2018.
\newblock {Sentencepiece: A Simple and Language Independent Subword Tokenizer and Detokenizer for Neural Text Processing}.
\newblock \emph{arXiv preprint arXiv:1808.06226}.

\bibitem[{Kulal et~al.(2019)Kulal, Pasupat, Chandra, Lee, Padon, Aiken, and Liang}]{kulal2019spoc}
Sumith Kulal, Panupong Pasupat, Kartik Chandra, Mina Lee, Oded Padon, Alex Aiken, and Percy Liang. 2019.
\newblock \href {https://arxiv.org/abs/1906.04908} {Spoc: Search-based pseudocode to code}.
\newblock \emph{Preprint}, arXiv:1906.04908.

\bibitem[{Labrak et~al.(2024)Labrak, Bazoge, Morin, Gourraud, Rouvier, and Dufour}]{labrak2024biomistral}
Yanis Labrak, Adrien Bazoge, Emmanuel Morin, Pierre-Antoine Gourraud, Mickael Rouvier, and Richard Dufour. 2024.
\newblock \href {https://arxiv.org/abs/2402.10373} {Biomistral: A collection of open-source pretrained large language models for medical domains}.
\newblock \emph{Preprint}, arXiv:2402.10373.

\bibitem[{Lachaux et~al.(2020)Lachaux, Roziere, Chanussot, and Lample}]{lachaux2020unsupervised}
Marie-Anne Lachaux, Baptiste Roziere, Lowik Chanussot, and Guillaume Lample. 2020.
\newblock \href {https://arxiv.org/abs/2006.03511} {Unsupervised translation of programming languages}.
\newblock \emph{Preprint}, arXiv:2006.03511.

\bibitem[{Lesort et~al.(2022)Lesort, Caccia, and Rish}]{lesort2022understanding}
Timothée Lesort, Massimo Caccia, and Irina Rish. 2022.
\newblock \href {https://arxiv.org/abs/2104.01678} {Understanding continual learning settings with data distribution drift analysis}.
\newblock \emph{Preprint}, arXiv:2104.01678.

\bibitem[{Lin et~al.(2024)Lin, Gou, Gong, Liu, Shen, Xu, Lin, Yang, Jiao, Duan, and Chen}]{lin2024rho1}
Zhenghao Lin, Zhibin Gou, Yeyun Gong, Xiao Liu, Yelong Shen, Ruochen Xu, Chen Lin, Yujiu Yang, Jian Jiao, Nan Duan, and Weizhu Chen. 2024.
\newblock \href {https://arxiv.org/abs/2404.07965} {Rho-1: Not all tokens are what you need}.
\newblock \emph{Preprint}, arXiv:2404.07965.

\bibitem[{Loshchilov and Hutter(2019)}]{loshchilov2019decoupled}
Ilya Loshchilov and Frank Hutter. 2019.
\newblock \href {https://arxiv.org/abs/1711.05101} {Decoupled weight decay regularization}.
\newblock \emph{Preprint}, arXiv:1711.05101.

\bibitem[{Loureiro et~al.(2022)Loureiro, Barbieri, Neves, Anke, and Camacho-Collados}]{loureiro2022timelms}
Daniel Loureiro, Francesco Barbieri, Leonardo Neves, Luis~Espinosa Anke, and Jose Camacho-Collados. 2022.
\newblock \href {https://arxiv.org/abs/2202.03829} {Timelms: Diachronic language models from twitter}.
\newblock \emph{Preprint}, arXiv:2202.03829.

\bibitem[{Ma et~al.(2023)Ma, Huang, Huang, Wang, Li, Zheng, Xie, Huang, and Jiang}]{ma2023ecomgptct}
Shirong Ma, Shen Huang, Shulin Huang, Xiaobin Wang, Yangning Li, Hai-Tao Zheng, Pengjun Xie, Fei Huang, and Yong Jiang. 2023.
\newblock \href {https://arxiv.org/abs/2312.15696} {Ecomgpt-ct: Continual pre-training of e-commerce large language models with semi-structured data}.
\newblock \emph{Preprint}, arXiv:2312.15696.

\bibitem[{OpenAI(2024)}]{openai2024gpt4}
OpenAI. 2024.
\newblock \href {https://arxiv.org/abs/2303.08774} {Gpt-4 technical report}.
\newblock \emph{Preprint}, arXiv:2303.08774.

\bibitem[{Parmar et~al.(2024)Parmar, Prabhumoye, Jennings, Patwary, Subramanian, Su, Zhu, Narayanan, Jhunjhunwala, Dattagupta, Jawa, Liu, Mahabaleshwarkar, Nitski, Brundyn, Maki, Martinez, You, Kamalu, LeGresley, Fridman, Casper, Aithal, Kuchaiev, Shoeybi, Cohen, and Catanzaro}]{parmar2024nemotron4}
Jupinder Parmar, Shrimai Prabhumoye, Joseph Jennings, Mostofa Patwary, Sandeep Subramanian, Dan Su, Chen Zhu, Deepak Narayanan, Aastha Jhunjhunwala, Ayush Dattagupta, Vibhu Jawa, Jiwei Liu, Ameya Mahabaleshwarkar, Osvald Nitski, Annika Brundyn, James Maki, Miguel Martinez, Jiaxuan You, John Kamalu, Patrick LeGresley, Denys Fridman, Jared Casper, Ashwath Aithal, Oleksii Kuchaiev, Mohammad Shoeybi, Jonathan Cohen, and Bryan Catanzaro. 2024.
\newblock \href {https://arxiv.org/abs/2402.16819} {Nemotron-4 15b technical report}.
\newblock \emph{Preprint}, arXiv:2402.16819.

\bibitem[{Qin et~al.(2022)Qin, Zhang, Lin, Liu, Li, Sun, and Zhou}]{qin2022elle}
Yujia Qin, Jiajie Zhang, Yankai Lin, Zhiyuan Liu, Peng Li, Maosong Sun, and Jie Zhou. 2022.
\newblock \href {https://arxiv.org/abs/2203.06311} {Elle: Efficient lifelong pre-training for emerging data}.
\newblock \emph{Preprint}, arXiv:2203.06311.

\bibitem[{Robins(1995)}]{Robins1995CatastrophicFR}
Anthony~V. Robins. 1995.
\newblock \href {https://api.semanticscholar.org/CorpusID:22882861} {Catastrophic forgetting, rehearsal and pseudorehearsal}.
\newblock \emph{Connect. Sci.}, 7:123--146.

\bibitem[{Rolnick et~al.(2019)Rolnick, Ahuja, Schwarz, Lillicrap, and Wayne}]{rolnick2019experience}
David Rolnick, Arun Ahuja, Jonathan Schwarz, Timothy~P. Lillicrap, and Greg Wayne. 2019.
\newblock \href {https://arxiv.org/abs/1811.11682} {Experience replay for continual learning}.
\newblock \emph{Preprint}, arXiv:1811.11682.

\bibitem[{Schwenk et~al.(2019)Schwenk, Wenzek, Edunov, Grave, and Joulin}]{schwenk2019ccmatrix}
Holger Schwenk, Guillaume Wenzek, Sergey Edunov, Edouard Grave, and Armand Joulin. 2019.
\newblock Ccmatrix: Mining billions of high-quality parallel sentences on the web.
\newblock \emph{arXiv preprint arXiv:1911.04944}.

\bibitem[{Scialom et~al.(2022)Scialom, Chakrabarty, and Muresan}]{scialom2022finetuned}
Thomas Scialom, Tuhin Chakrabarty, and Smaranda Muresan. 2022.
\newblock \href {https://arxiv.org/abs/2205.12393} {Fine-tuned language models are continual learners}.
\newblock \emph{Preprint}, arXiv:2205.12393.

\bibitem[{Shi et~al.(2022)Shi, Suzgun, Freitag, Wang, Srivats, Vosoughi, Chung, Tay, Ruder, Zhou, Das, and Wei}]{shi2022language}
Freda Shi, Mirac Suzgun, Markus Freitag, Xuezhi Wang, Suraj Srivats, Soroush Vosoughi, Hyung~Won Chung, Yi~Tay, Sebastian Ruder, Denny Zhou, Dipanjan Das, and Jason Wei. 2022.
\newblock \href {https://arxiv.org/abs/2210.03057} {Language models are multilingual chain-of-thought reasoners}.
\newblock \emph{Preprint}, arXiv:2210.03057.

\bibitem[{Soviany et~al.(2022)Soviany, Ionescu, Rota, and Sebe}]{soviany2022curriculum}
Petru Soviany, Radu~Tudor Ionescu, Paolo Rota, and Nicu Sebe. 2022.
\newblock \href {https://arxiv.org/abs/2101.10382} {Curriculum learning: A survey}.
\newblock \emph{Preprint}, arXiv:2101.10382.

\bibitem[{Su et~al.(2023)Su, Lu, Pan, Murtadha, Wen, and Liu}]{su2023roformer}
Jianlin Su, Yu~Lu, Shengfeng Pan, Ahmed Murtadha, Bo~Wen, and Yunfeng Liu. 2023.
\newblock \href {https://arxiv.org/abs/2104.09864} {Roformer: Enhanced transformer with rotary position embedding}.
\newblock \emph{Preprint}, arXiv:2104.09864.

\bibitem[{Team(2024)}]{geminiteam2024gemini}
Gemini Team. 2024.
\newblock \href {https://arxiv.org/abs/2312.11805} {Gemini: A family of highly capable multimodal models}.
\newblock \emph{Preprint}, arXiv:2312.11805.

\bibitem[{Team et~al.(2024)Team, Ormazabal, Zheng, de~Masson~d'Autume, Yogatama, Fu, Ong, Chen, Lamprecht, Pham, Ong, Aleksiev, Li, Henderson, Bain, Artetxe, Relan, Padlewski, Liu, Chen, Phua, Yang, Tay, Wang, Zhu, and Xie}]{rekateam2024reka}
Reka Team, Aitor Ormazabal, Che Zheng, Cyprien de~Masson~d'Autume, Dani Yogatama, Deyu Fu, Donovan Ong, Eric Chen, Eugenie Lamprecht, Hai Pham, Isaac Ong, Kaloyan Aleksiev, Lei Li, Matthew Henderson, Max Bain, Mikel Artetxe, Nishant Relan, Piotr Padlewski, Qi~Liu, Ren Chen, Samuel Phua, Yazheng Yang, Yi~Tay, Yuqi Wang, Zhongkai Zhu, and Zhihui Xie. 2024.
\newblock \href {https://arxiv.org/abs/2404.12387} {Reka core, flash, and edge: A series of powerful multimodal language models}.
\newblock \emph{Preprint}, arXiv:2404.12387.

\bibitem[{Touvron et~al.(2023)Touvron, Martin, Stone, Albert, Almahairi, Babaei, Bashlykov, Batra, Bhargava, Bhosale et~al.}]{touvron2023llama2}
Hugo Touvron, Louis Martin, Kevin Stone, Peter Albert, Amjad Almahairi, Yasmine Babaei, Nikolay Bashlykov, Soumya Batra, Prajjwal Bhargava, Shruti Bhosale, et~al. 2023.
\newblock {Llama 2: Open Foundation and Fine-tuned Chat Models}.
\newblock \emph{arXiv preprint arXiv:2307.09288}.

\bibitem[{Vaswani et~al.(2017)Vaswani, Shazeer, Parmar, Uszkoreit, Jones, Gomez, Kaiser, and Polosukhin}]{Vaswani+2017}
Ashish Vaswani, Noam Shazeer, Niki Parmar, Jakob Uszkoreit, Llion Jones, Aidan~N Gomez, \L~ukasz Kaiser, and Illia Polosukhin. 2017.
\newblock \href {https://proceedings.neurips.cc/paper_files/paper/2017/file/3f5ee243547dee91fbd053c1c4a845aa-Paper.pdf} {Attention is all you need}.
\newblock In \emph{Advances in Neural Information Processing Systems}, volume~30. Curran Associates, Inc.

\bibitem[{Wang et~al.(2022)Wang, Yang, Huang, Jiao, Yang, Jiang, Majumder, and Wei}]{wang2022text}
Liang Wang, Nan Yang, Xiaolong Huang, Binxing Jiao, Linjun Yang, Daxin Jiang, Rangan Majumder, and Furu Wei. 2022.
\newblock Text embeddings by weakly-supervised contrastive pre-training.
\newblock \emph{arXiv preprint arXiv:2212.03533}.

\bibitem[{Winata et~al.(2023)Winata, Xie, Radhakrishnan, Wu, Jin, Cheng, Kulkarni, and Preotiuc-Pietro}]{winata2023overcoming}
Genta~Indra Winata, Lingjue Xie, Karthik Radhakrishnan, Shijie Wu, Xisen Jin, Pengxiang Cheng, Mayank Kulkarni, and Daniel Preotiuc-Pietro. 2023.
\newblock \href {https://arxiv.org/abs/2305.16252} {Overcoming catastrophic forgetting in massively multilingual continual learning}.
\newblock \emph{Preprint}, arXiv:2305.16252.

\bibitem[{Wu et~al.(2024)Wu, Gan, Ge, Lu, Wang, Feng, Shan, and Luo}]{wu2024llama}
Chengyue Wu, Yukang Gan, Yixiao Ge, Zeyu Lu, Jiahao Wang, Ye~Feng, Ying Shan, and Ping Luo. 2024.
\newblock \href {https://arxiv.org/abs/2401.02415} {Llama pro: Progressive llama with block expansion}.
\newblock \emph{Preprint}, arXiv:2401.02415.

\bibitem[{Yadav et~al.(2023)Yadav, Sun, Ding, Li, Zhang, Tan, Ma, Bhatia, Nallapati, Ramanathan, Bansal, and Xiang}]{yadav2023exploring}
Prateek Yadav, Qing Sun, Hantian Ding, Xiaopeng Li, Dejiao Zhang, Ming Tan, Xiaofei Ma, Parminder Bhatia, Ramesh Nallapati, Murali~Krishna Ramanathan, Mohit Bansal, and Bing Xiang. 2023.
\newblock \href {https://arxiv.org/abs/2307.02435} {Exploring continual learning for code generation models}.
\newblock \emph{Preprint}, arXiv:2307.02435.

\bibitem[{Yang et~al.(2024)Yang, Gao, Xue, and Alexandersson}]{yang2024pllama}
Xianjun Yang, Junfeng Gao, Wenxin Xue, and Erik Alexandersson. 2024.
\newblock \href {https://arxiv.org/abs/2401.01600} {Pllama: An open-source large language model for plant science}.
\newblock \emph{Preprint}, arXiv:2401.01600.

\bibitem[{Zan et~al.(2022)Zan, Chen, Yang, Lin, Kim, Guan, Wang, Chen, and Lou}]{zan2022cert}
Daoguang Zan, Bei Chen, Dejian Yang, Zeqi Lin, Minsu Kim, Bei Guan, Yongji Wang, Weizhu Chen, and Jian-Guang Lou. 2022.
\newblock \href {https://arxiv.org/abs/2206.06888} {Cert: Continual pre-training on sketches for library-oriented code generation}.
\newblock \emph{Preprint}, arXiv:2206.06888.

\bibitem[{Zellers et~al.(2019)Zellers, Holtzman, Bisk, Farhadi, and Choi}]{Zellers2019HellaSwagCA}
Rowan Zellers, Ari Holtzman, Yonatan Bisk, Ali Farhadi, and Yejin Choi. 2019.
\newblock Hellaswag: Can a machine really finish your sentence?
\newblock In \emph{ACL}.

\bibitem[{Çağatay Yıldız et~al.(2024)Çağatay Yıldız, Ravichandran, Punia, Bethge, and Ermis}]{yıldız2024investigating}
Çağatay Yıldız, Nishaanth~Kanna Ravichandran, Prishruit Punia, Matthias Bethge, and Beyza Ermis. 2024.
\newblock \href {https://arxiv.org/abs/2402.17400} {Investigating continual pretraining in large language models: Insights and implications}.
\newblock \emph{Preprint}, arXiv:2402.17400.

\end{thebibliography}

\appendix

\section{Data}
\label{sec:data_sources}

\subsection{Multilingual Data}
\label{sec:appendix_data_sources_multilingual}

The 53 multilingual languages contained within the pretraining set are: AR, AZ, BG, BN, CA, CS, DA, DE, EL, ES, ET, FA, FI, FR, GL, HE, HI, HR, HU, HY, ID, IS, IT, JA, KA, KK, KN,	KO, LT, LV, MK, ML, MR, NE, NL, NO, PL, PT, RO, RU, SK, SL, SQ, SR, SV, TA, TE, TH, TR, UK, UR, VI, and ZH.

\subsection{Code Data}
\label{sec:appendix_data_sources_code}

The 43 programming languags contained within our pretraining set are: assembly, c, c-sharp, common-lisp, cpp, css, cuda, dart, dockerfile, fortran, go, haskell, html, java, javascript, json, julia, jupyter-scripts, lua, makefile, markdown, mathematica, omniverse, pascal, perl, php, python, R, restructuredtext, ruby, rust, scala, shell, sql, swift, systemverilog, tex, typescript, verilog, vhdl, visual-basic, xml, and yaml.

\section{Experiments}
\label{sec:appendix_experiments}

The evaluation results across all considered tasks are shared below for each of our experiments. 
\begin{table}[!h]
\centering
  \begin{tabular}{lc}
   \toprule
    \textbf{Task} & Pretrained Model \\
    \midrule
    MMLU & 59.3 \\
    HellaSwag & 80.4 \\
    HumanEval & 31.1 \\
    MGSM (ES, JA, TH) & 24.9 \\
    \bottomrule
  \end{tabular}
  \caption{Model accuracy after 8T tokens of pretraining. We find that the model struggles on STEM based reasoning tasks due to its low scores on MGSM and STEM substasks of MMLU.}
  \label{tab:base_model_results_detailed}
\end{table}

\subsection{Data Distribution}
\label{sec:appendix_experiments_data_distributions}

Table \ref{tab:data_distr_exps_results_detailed} shares the results across all tasks for each experiment mentioned within Section \ref{sec:data_distributions}.

\begin{table*}[!h]
\centering
  \begin{tabular}{lcccc}
   \toprule
    \textbf{Data Blend} & \textbf{MMLU} & \textbf{HellaSwag} & \textbf{HumanEval} & \textbf{MGSM (ES, JA, TH)} \\
    \midrule
    Pretraining &  61.9 & 81.2  & 28.1  & 34.7  \\
    QA &  62 & 78.7  & 32.9 & 40.1  \\
    Pretraining (250B) + QA (50B)  &  62.6 & 82.2 & 29.9 & 42.4  \\
    \midrule
    Pretraining &  61.9 & 81.2  & 28.1  & 34.7  \\
    Reweight Domains &  61.9 & 81.7  & 29.9 & 33.2  \\
    Pretraining w/ High Quality Web  &  62.2 & 80.9 & 34.1 & 32.9  \\
    No Web &  62.3 & 81.8  & 29.9 & 37.7  \\
    Upweight Non Web w/ High Quality Web  &  62.6 & 81.4 & 31.7 & 32.1  \\
    \midrule
    QA 1 &  63.0 & 82.4  & 29.9  & 41.9  \\
    QA 2 (+STEM, +World Knowledge) &  63.9 & 82.3  & 29.3 & 36.7  \\
    QA 3 (+STEM, +Chat)  &  64.1 & 82.2 & 28.7 & 44.7  \\
    \midrule
    QA &  64.2 & 82.4 & 30.5 & 44.5 \\
    QA w/ Upweighted STEM  & 64.1   & 82.3  & 28.1  & 42.9  \\
    QA w/ 1.5e QA data &  64.1 & 82.2 & 28.7 & 44.7  \\
    QA w/ 3.5e QA data   & 64.4   & 27.4  & 82.4  & 43.3  \\
    \bottomrule
  \end{tabular}
  \caption{Per-task evaluation results of each experiment mentioned within Section \ref{sec:data_distributions} on defining data distributions for continued pretraining.}
  \label{tab:data_distr_exps_results_detailed}
\end{table*}

\subsection{Learning Rate Schedule}
\label{sec:appendix_experiments_learning_rates}

\begin{table*}[!h]
\centering
  \begin{tabular}{lcccc}
   \toprule
    \textbf{LR Schedule} & \textbf{MMLU} & \textbf{HellaSwag} & \textbf{HumanEval} & \textbf{MGSM (ES, JA, TH)} \\
    \midrule
    Decay to $\frac{\eta_{max_{\text{ct}}}}{10}$ &  63.9 & 82.4 & 29.3 & 43.7  \\
    Decay to $\frac{\eta_{max_{\text{ct}}}}{100}$ &   64.2 & 82.2 & 31.1 & 45.2 \\
    Decay to 0  &  64.2 & 30.5 & 82.4 & 44.5  \\
    \bottomrule
  \end{tabular}
  \caption{Per-task evaluation results of the experiments mentioned in Table \ref{tab:lr_decay_results} on identifying an appropriate learning rate decay schedule for continued pretraining. }
  \label{tab:lr_decay_results_detailed}
\end{table*}

\begin{figure}[h]
    \centering
    \includegraphics[width=\linewidth]{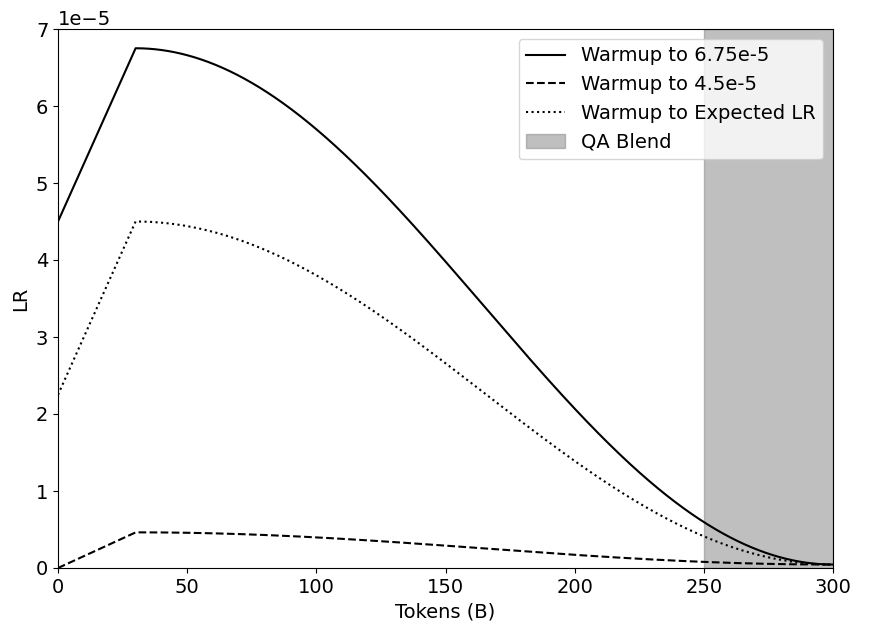}
    \caption{Cosine decay schedule with the various levels of warmup which we experiment with.}
    \label{fig:warmup_lrs}
\end{figure}

In identifying a learning rate schedule for continued pretraining, we experiment with various degrees of warmup and values of $\eta_{max_{\text{ct}}}$. The combinations we consider are: warmup from $\eta_{min}$ to $\eta_{max_{\text{ct}}} = 1.5*\eta_{min}$,  warmup from $0.5*\eta_{min}$ to $\eta_{max_{\text{ct}}} = \eta_{min}$, and warmup from 0 to what the expected learning rate value would be had the pretraining learning rate schedule been extended to incorporate the continued training tokens (i.e., from 8T to 8.3T). We use $\eta_{min}$ to specify the minimum learning rate value of the pretrained model, which is $4.5e\text{-}5$. Figure \ref{fig:warmup_lrs} highlights each of these schedules, and we note that these combinations were chosen to quantify different degrees of aggressiveness when using warmup in a continued pretraining learning rate schedule. 

As highlighted in Table \ref{tab:warmup_results_detailed}, we find that including any level of warmup within the continued training learning rate schedule causes regressions in evaluation accuracies, indicating that it is best to decay directly from $\eta_{min}$.




\begin{table*}[!h]
\centering
  \begin{tabular}{lccccc}
   \toprule
    \textbf{LR Schedule} & \textbf{MMLU} & \textbf{HellaSwag} & \textbf{HumanEval} & \textbf{MGSM (ES, JA, TH)} & \textbf{Avg. Acc.}  \\
    \midrule
    Warmup to $6.75e\text{-}5$ &  64.0 & 81.9 & 31.1 & 42.3 & 54.8  \\
    Warmup to $4.5e\text{-}5$ &   64.0 & 82.1 & 32.9 & 41.5 & 55.1 \\
    Warmup to Expected LR  &  63.3 & 82.1 & 31.7 & 42.5 & 54.9 \\
    No Warmup &   64.2 & 31.1 & 82.2 & 45.2 &  \textbf{55.7} \\
    \bottomrule
  \end{tabular}
  \caption{Comparison of including warmup within learning rate schedules for continued pretraining. No warmup achieves the best evaluation results. }
  \label{tab:warmup_results_detailed}
\end{table*}

In addition to cosine annealing, we experiment with the WSD learning rate scheduler \citep{hu2024minicpm}. Table \ref{tab:lr_decay_results_detailed_wsd} compares the best found setting of WSD with cosine annealing. The WSD schedule produces significantly lower evaluation accuracies than cosine annealing. We hypothesize that in continued pretraining, switching the decay schedule from the one used during pretraining is harmful. Hence, for models pretrained with cosine annealing, the learning rate schedule in continued training should also use cosine annealing.

\begin{table*}[!h]
\centering
  \begin{tabular}{lccccc}
   \toprule
    \textbf{LR Schedule} & \textbf{MMLU} & \textbf{HellaSwag} & \textbf{HumanEval} & \textbf{MGSM (ES, JA, TH)} & \textbf{Avg. Acc.}   \\
    \midrule
    WSD  &  63.6 & 80.2 & 28.1 & 39.5  & 52.8 \\
    Cosine Annealing  &   64.2 & 82.2 & 31.1 & 45.2 & \textbf{55.7} \\
    \bottomrule
  \end{tabular}
  \caption{We find that WSD causes significant regression in evaluation accuracy compared to cosine annealing. Both learning rate schedules were decayed till $\frac{\eta_{max_{\text{ct}}}}{100}$.}
  \label{tab:lr_decay_results_detailed_wsd}
\end{table*}



\subsection{Switch of Data Distributions}
\label{sec:appendix_experiments_distribution_switch}

Table \ref{tab:distribution_switch_100B} highlights that the findings of our experiments in Section \ref{sec:experiments_distribution_switch} also hold at the continued training token horizon of 100B tokens. This indicates that regardless of the number of continued training tokens, transitioning between the GB and QB distributions at $\frac{\eta_{max_{\text{ct}}}}{5}$ is optimal.

\begin{table*}[!h]
\centering
  \begin{tabular}{lcccc}
   \toprule
    \textbf{Distribution Switch} & \textbf{MMLU} & \textbf{HellaSwag} & \textbf{HumanEval} & \textbf{MGSM (ES, JA, TH)} \\
    \midrule
    At $\eta_{max_{\text{ct}}}$ (from step 0) &  65.0 & 78.7 & 29.9 & 37.7 \\
    At $\frac{\eta_{max_{\text{ct}}}}{2}$  &  60.9 & 81.6 & 32.3 & 44.1 \\
    At $\frac{\eta_{max_{\text{ct}}}}{5}$  &  63.8 & 82.2 & 32.3 & 46.1 \\
    At $\frac{\eta_{max_{\text{ct}}}}{10}$  &  63.9 & 82.2 & 29.3 & 44.7 \\
    At $\frac{\eta_{max_{\text{ct}}}}{50}$  &  63.3 & 81.6 & 31.1 & 42.3 \\
    \bottomrule
  \end{tabular}
  \caption{Per-task evaluation results of the experiments mentioned in Table \ref{tab:distribution_switch} on how to switch between data distributions in continued pretraining.}
  \label{tab:distribution_switch_results_detailed}
\end{table*}

\begin{table*}[!h]
\centering
  \begin{tabular}{lccccc}
   \toprule
    \textbf{Distribution Switch} & \textbf{MMLU} & \textbf{HellaSwag} & \textbf{HumanEval} & \textbf{MGSM (ES, JA, TH)} & \textbf{AVG} \\
    \midrule
    At $\eta_{max_{\text{ct}}}$ (from step 0) &  64.1 & 79.2 & 31.1 & 40.0 & 53.6  \\
    At $\frac{\eta_{max_{\text{ct}}}}{2}$  &  63.2 & 81.6 & 27.4 & 44.1 & 54.1  \\
    At $\frac{\eta_{max_{\text{ct}}}}{5}$  &  63.0 & 81.9 & 31.7 & 43.6 & \textbf{55.0}  \\
    At $\frac{\eta_{max_{\text{ct}}}}{10}$ &  63.6 & 81.8 & 30.5 & 39.7 & 53.9  \\
    At $\frac{\eta_{max_{\text{ct}}}}{50}$ &  63.3 & 81.6 & 31.1 & 42.3 & 54.6  \\
    \bottomrule
  \end{tabular}
  \caption{Ablation of the data distribution switch experiments at a continued pretraining scale of 100B tokens. As found for the 300B token continued training horizon, switching distributions at $\frac{\eta_{max_{\text{ct}}}}{5}$ achieves the highest accuracy.  }
  \label{tab:distribution_switch_100B}
\end{table*}

\section{Ablations}
\subsection{Varying Token Horizons}
\label{sec:appendix_continued_training_varying_amounts}

\begin{figure*}[h]
    \centering
    \includegraphics[width=\linewidth]{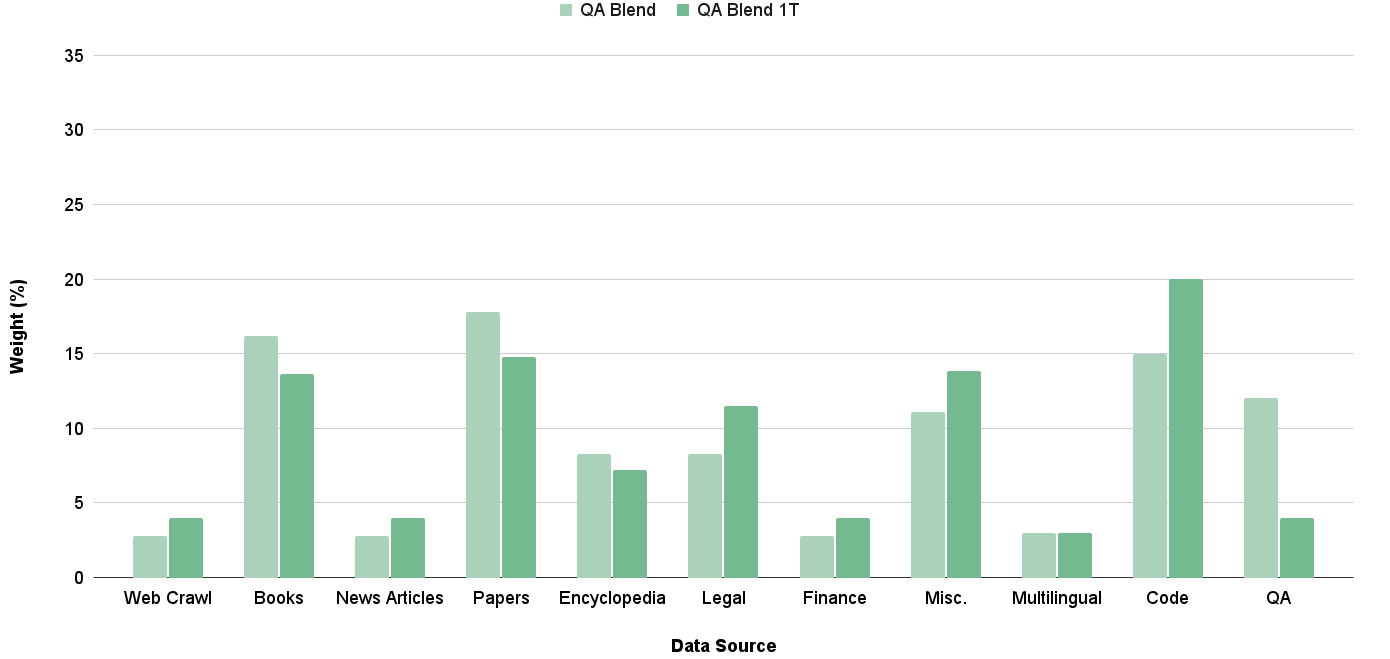}
    \caption{Distribution of the QB blend when extending the number of continued pretraining tokens to 1T.}
    \label{fig:qb_distr_length}
\end{figure*}

When extending the number of continued pretraining tokens to 1T, we found that our existing QB distribution would cause the small QA dataset to be trained on for a large number of epochs. To correct for this, we reduce the weight on the QA datset so that it would be trained on for no more than 4 epochs. Figure \ref{fig:qb_distr_length} demonstrates the distribution of the QB when used at the scale of 1T continued pretraining tokens. 

\begin{table*}[!h]
\centering
  \begin{tabular}{lccccc}
   \toprule
    \textbf{Num CT Tokens} & \textbf{MMLU} & \textbf{HellaSwag} & \textbf{HumanEval} & \textbf{MGSM (ES, JA, TH)} & \textbf{AVG} \\
    \midrule
    0B & 59.3 & 80.4 & 31.1  & 24.9 & 48.9  \\
   100B  &  63.0 & 81.9 & 31.7 & 43.6 & 55.0  \\
    300B  &  63.8 & 82.2 & 32.3 & 46.1 & 56.1 \\
    1T  &  65.3 & 82.4 & 34.1 & 45.5 \\
    \bottomrule
  \end{tabular}
  \caption{Per-task evaluation results of the experiments mentioned in Table \ref{ct_mined_documents_results} on how the identified continued pretraining recipe performs at varying amounts of continued training tokens.}
  \label{tab:ct_varying_tokens_results_detailed}
\end{table*}

\begin{table*}[!h]
\centering
  \begin{tabular}{lcccc}
   \toprule
    \textbf{Blend} & \textbf{MMLU} & \textbf{HellaSwag} & \textbf{HumanEval} & \textbf{MGSM (ES, JA, TH)} \\
    \midrule
    CT 1T  &  65.3 & 82.4 & 34.1 & 45.5 \\
    CT 1T w/ Mined Docs &  66.6  &  81.7 & 36.6 & 46.7  \\
    \bottomrule
  \end{tabular}
  \caption{Per-task evaluation results of the experiments mentioned in Table \ref{ct_mined_documents_results} on how document mining increases the utility of existing data sources in continued pretraining.}
  \label{tab:ct_mined_documents_results_detailed}
\end{table*}

\end{document}